\pdfoutput=1

\documentclass[11pt]{article}

\usepackage[]{emnlp2021}

\usepackage{times}
\usepackage{latexsym}
\usepackage{todonotes}
\usepackage{amsmath}
\usepackage{amsfonts}
\usepackage{caption}
\usepackage{subcaption}
\usepackage{booktabs}
\usepackage{multirow}
\usepackage{comment} 
\usepackage{hyperref}
\usepackage[detect-none]{siunitx}
\newcommand{\citepos}[1]{\citeauthor{#1}'s (\citeyear{#1})}
\usepackage[T1]{fontenc}

\usepackage[utf8]{inputenc}

\usepackage{microtype}

\title{Can Language Models Encode Perceptual Structure Without Grounding? \\A Case Study in Color}

\author{Mostafa Abdou\thanks{For correspondence: \texttt{\{abdou,soegaard\}@di.ku.dk}} \\
  University of Copenhagen \\
   \\\And
  Artur Kulmizev \\
  Uppsala University \\
  \\ \And
  Daniel Hershcovich \\
  University of Copenhagen \\
   \\ \AND
  Stella Frank \\
  University of Trento \\
   \\ \And
  Ellie Pavlick \\
  Brown University \\
   \\ \And
  Anders Søgaard \\
  University of Copenhagen \\
  
  }

\begin{document}
\maketitle
\begin{abstract}

Pretrained language models have been shown to encode relational information, such as the relations between entities or concepts in knowledge-bases --- (Paris, Capital, France). However, simple relations of this type can often be recovered heuristically and the extent to which models implicitly reflect topological structure that is grounded in world, such as perceptual structure, is unknown. To explore this question, we conduct a thorough case study on color. Namely, we employ a dataset of monolexemic color terms and color chips represented in CIELAB, a color space with a perceptually meaningful distance metric.

Using two methods of evaluating the structural alignment of colors in this space with text-derived color term representations, we find significant correspondence. Analyzing the differences in alignment across the color spectrum, we find that warmer colors are, on average, better aligned to the perceptual color space than cooler ones, suggesting an intriguing connection to findings from recent work on efficient communication in color naming. Further analysis suggests that differences in alignment are, in part, mediated by collocationality and differences in syntactic usage, posing questions as to the relationship between color perception and usage and context. 

\end{abstract}

\section{Introduction}

\begin{figure}[t!]
\centering
\includegraphics[scale=0.45]{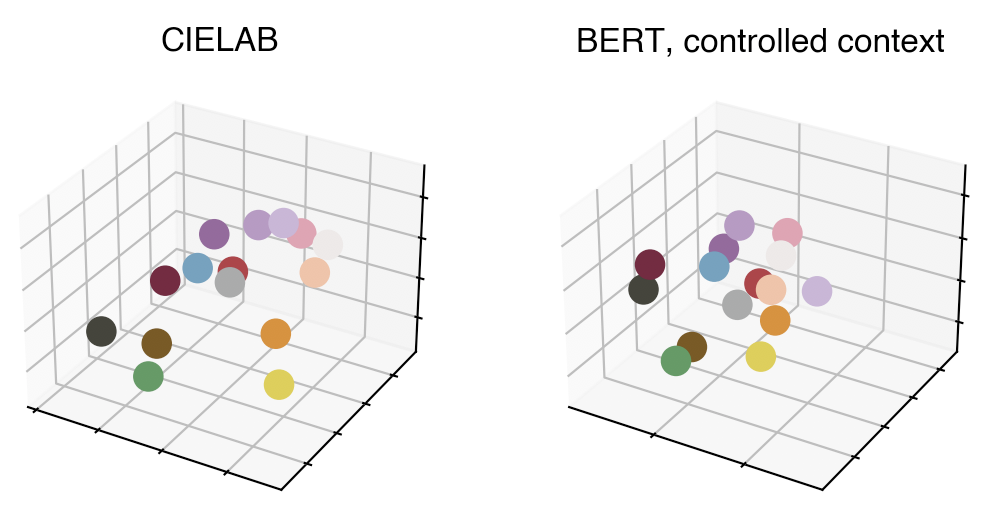}
\caption{Right: Color orientation in 3d CIELAB space. Left: linear mapping from BERT (CC, see \S\ref{sec:method}) color term embeddings to the CIELAB space. }
\label{fig:first}
\end{figure}

Without grounding or interaction with the world, language models (LMs) learn representations that encode various aspects of formal linguistic structure (e.g., morphosyntax \cite{tenney-etal-2019-bert}) and semantic information (e.g., lexical similarity \cite{NEURIPS2019_159c1ffe}). Beyond this, it has been suggested that text-only training data is enough for LMs to also acquire factual and relational information about the world  \cite{davison-etal-2019-commonsense,petroni-etal-2019-language}. This includes, for instance, some features of concrete and abstract concepts, such as objects' attributes and affordances \cite{forbes-etal-2019-neural,weir2020probing}. Furthermore, the representational geometry of LMs has been found to naturally reflect human lexical similarity and relatedness judgements, as well as analogy relationships \cite{chronis-erk-2020-bishop}. However, the extent to which these models reflect the structures that exist in humans' perceptual world---such as the topology of visual perception \cite{chen1982topological}, the structure of the color spectrum \cite{ennis2019geometrical, provenzi2020geometry}, or of odour spaces \cite{rossiter1996structure, chastrette1997trends}---is not well-understood. 

If LMs are indeed able to capture such topologies---in some domains, at least---it would mean that these structures are a) somehow reflected in language and, thereby, encoded in the textual training data on which models are trained, and b) learnable using models' current training objectives and architectural inductive biases. To the extent they are not, the question becomes whether the information is not there in the data, or whether model and training objective limitations are to blame. Certainly, this latter point relates to an ongoing debate regarding what exactly language models can be expected to learn from ungrounded form alone \cite{bender-koller-2020-climbing, bisk2020experience, merrill2021provable}. While there have been many interesting theoretical debates around this topic, few studies have tried to address this question empirically.

\begin{figure*}[ht]
\centering
\includegraphics[width=0.99999\textwidth]{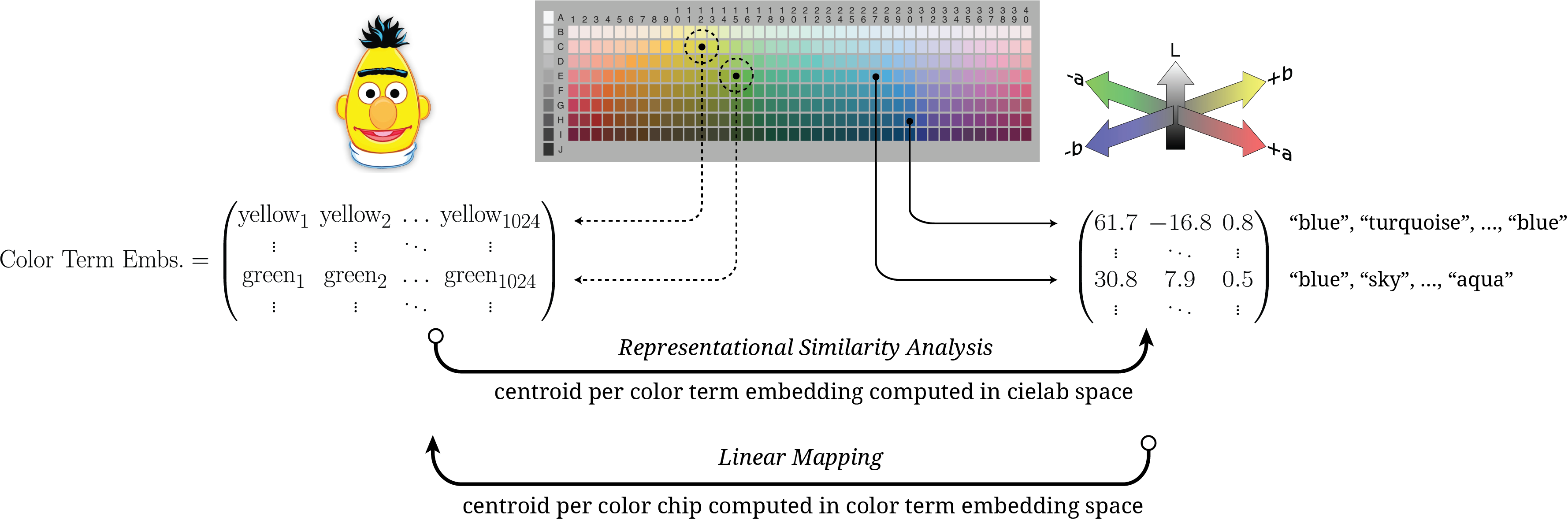}
\caption{Our experimental setup. In the center is a Munsell color chart. Each chip in the chart is represented in the CIELAB space (right) and has 51 color term annotations. Color term embeddings are extracted through various methods. In the Representation Similarity Analysis experiments, a corresponding color chip centroid is computed in the CIELAB space. In the Linear Mapping experiments, a color term embedding centroid is computed per chip. }
\label{fig:main}
\end{figure*}

In this paper, we conduct a case study on color. Indeed, color perception in humans and its relation to speakers' use of color terms has long been the subject of studies in cognitive science \cite{kay1978linguistic, berlin1991basic, regier2007color, kay2009world}. To this end, spaces have been defined in which Euclidean distances between related
colors are correlated with reported perceptual differences.\footnote{The differences between color stimuli which are perceived by human observers.} In addition, the semantics of color terms have long been understood to hold particular linguistic significance, as they are theorised to be subject to universal constraints that arise directly from the neurophysiological mechanisms and properties underlying visual perception and cognition \cite{kay1978linguistic,berlin1991basic, kay1991biocultural}.\footnote{These theories have been contested by work arguing for linguistic relativism (cf. the \textit{Sapir--Whorf Hypothesis}), which emphasizes the arbitrariness of language and the relativity of semantic structures and minimizes the role of universals. Such critiques have, however, been accommodated for in the Berlin \& Kay paradigm \cite{berlin1991basic}, the basic assumptions of which, such as the existence of at least some perceptually-determined universal constraints on color naming, remain widely accepted.} Due to these factors, color offers a useful test-bed for investigating whether or not structural information about the topology of the perceptual world might be encoded in linguistic representations. 

To explore this in detail, we employ a dataset of English color terms and their corresponding color chips\footnote{Each chip is a unique color sample from the Munsell chart, which is made up of $330$ such samples which cover the space of colors perceived by humans. See \S\ref{sec:method}.}, the latter of which are represented in CIELAB --- a perceptually uniform color space. In addition to the color chip CIELAB coordinates, we extract linguistic representations for the corresponding color terms. With these two representations in mind (see Figure~\ref{fig:first} for a demonstrative plot from our experiments), we employ two methods of measuring structural correspondence, with which we evaluate the alignment between the two spaces. Figure~\ref{fig:main} shows an illustration of the experimental setup. We find that the structures of various language model representations show alignment with the structure of the CIELAB space, demonstrating that some approximation of perceptual color space topology can indeed be learned from text alone. We also show that part of this distributional signal is learnable by simple models --- e.g. models based on pointwise mutual information (PMI) statistics --- although large-scale language model pretraining (e.g., BERT) encodes the topology markedly better. 


Analysis shows that larger language models align better than smaller ones and that much of the variance in CIELAB space can be explained by low-dimensional subspaces of LM-induced color term representations. To better understand the results, we also analyse the differences in alignment across the color spectrum, observing that warm colors are generally better aligned than cool ones. Further investigation reveals a connection to findings reported in work on communication efficiency in color naming, which posits that warmer colors are communicated more efficiently. Finally, we investigate various corpus statistics which could influence alignment, finding that a measure of color term collocationality based on PMI statistics corresponds to lower alignment, while the entropy of a color term's dependency relation distribution (i.e. terms occurring as adjectival modifiers, nominal subjects, etc.) and how often it occurs as an adjectival modifier correspond to a stronger one.

\section{Methodology}
\label{sec:method}
\paragraph{Color data}
We employ the Color Lexicon of American English, which provides extensive data on color naming. The lexicon consists of $51$ monolexemic color name judgements for each of the $330$ Munsell Chart color chips\footnote{\url{http://www1.icsi.berkeley.edu/wcs/images/jrus-20100531/wcs-chart-4x.png}} \cite{lindsey2014color}. The color terms are solicited through a free-naming task, resulting in $122$ terms.

\paragraph{Perceptual color space} Following previous work \cite{regier2007color, zaslavsky2018efficient, chaabouni2021communicating}, we map colors to their corresponding points in the $3$D CIELAB space, where the first dimension $L$ expresses lightness, the second $A$ expresses position between red and green, and the third $B$ expresses the position between blue and yellow. Distances between colors in the space correspond to their perceptual difference. 

\paragraph{Language models}
Our analysis is conducted on three widely used language models (LMs): BERT \cite{devlin-etal-2019-bert} and  RoBERTa \cite{liu2019roberta}, both of which employ a masked language modelling objective, and ELECTRA \cite{clark2020electra}, which is trained instead with a discriminative token replacement detection objective.\footnote{\texttt{bert-large-uncased; roberta-large; electra-large-discriminator}}

\paragraph{Baselines} In addition to the aforementioned language models, we consider two different baselines:

\begin{itemize}
 \setlength{\itemsep}{0.1mm}
  \setlength{\parskip}{.1mm}
  \setlength{\parsep}{0mm}
    \item PMI statistics, which are computed\footnote{Using Hyperwords: \url{https://bitbucket.org/omerlevy/hyperwords}} for the color terms in common crawl, using window sizes of $1$ (pmi-1), $2$ (pmi-2), and $3$ (pmi-3). The result is a vocabulary length vector quantifying the likelihood of co-occurrence of the color term with every other vocabulary item in within that window.
    \item Word-type FastText embeddings trained on Common Crawl \cite{bojanowski2017enriching}. 
\end{itemize}

\paragraph{Representation Extraction}
We follow \newcite{bommasani-etal-2020-interpreting} and \newcite{vulic2020probing} in defining  configurations for the extraction of word-type representations from LM hidden states. In the first configuration (NC), a color term is encoded without context, with the appropriate delimiter tokens attached (e.g. \texttt{[CLS] red [SEP]} for BERT). In the second, $S$ sentential contexts that include the color term are encoded and the hidden states representing these contexts are mean pooled. These $S$ contexts are either randomly sampled from common crawl (RC), or deterministically generated to allow for control over contextual variation (CC). If a color term is split by an LM's tokenizer into more than one token, subword token encodings are averaged over. For each color term and configuration, an embedding vector of hidden state dimension $d_{LM}$ is extracted per layer, per model. 

\paragraph{Controlled context}
To control for the effect of variation in the sentence contexts used to construct color term representations, we employ a templative approach to generate a set of identical contexts for all color terms. When generating controlled contexts, we create three frames in which the terms can appear: 

\begin{itemize}
 \setlength{\itemsep}{0.1mm}
  \setlength{\parskip}{.1mm}
  \setlength{\parsep}{0mm}
    \item \textsc{Copula}: the \texttt{<obj>} is \texttt{<col>}
    \item \textsc{Possession}: i have a \texttt{<col> <obj>}
    \item \textsc{Spatial}: the \texttt{<col> <obj>} is there
\end{itemize}

We use these frames in order to limit the
contextual variation across colors (\texttt{<col>}) and to isolate their representations amidst as little semantic interference as possible, all while retaining a naturalistic quality to the input. We also aggregate over numerous object nouns (\texttt{<obj>}), which the color terms are used to describe. We select objects from the \newcite{mcrae2005semantic} data which are labelled in the latter as plausibly occurring in many colors and which are stratified across $13$ category sets, e.g. \textit{fan} $\in$ \texttt{APPLIANCES}, \textit{skirt} $\in$ \texttt{CLOTHING}, etc. Collapsing over categories, we generate sentences combinatorially across frames, objects and color terms, resulting in $3 \times 122 \times 18 = 6588$ sentences, $366$ per term.

\section{Evaluation}
\label{sec:eval}
 We employ two complimentary evaluation methods to gauge the correspondence of the color term text-derived representations to the perceptual color space. The first, Representation Similarity Analysis (RSA), is non-parametric and uses pairwise comparisons of stimuli to provide a measure of the global topological alignment between two spaces. The second employs a learned linear mapping, evaluating the extent to which two spaces can be aligned via transformation (rotation, scaling, etc.). 

\paragraph{RSA} \citep{kriegeskorte2008representational} is a method of relating different representational modalities, which was first employed in neuroscientific studies. RSA abstracts away from activity patterns themselves (e.g. neuron values in representational vectors) and instead computes representational (dis)-similarity matrices (RSMs), which characterize the information carried by a given representation method through global (dis)-similarity structure. Kendall's rank correlation coefficient ($\tau$) is computed between RSMs derived from the two spaces, providing a summary statistic indicative of the overall representational alignment between them. RSA is non-parametric and therefore circumvents many of the various methodological weaknesses associated with the probing paradigm \cite{belinkov2021probing}. 

For each color term, we compute a centroid in the CIELAB space following the approach described in \newcite{lindsey2014color}. Each centroid is defined as the average CIELAB coordinate of the samples (i.e. color chips) that were named with the corresponding term (across the 51 subjects). This results in $N$ parallel points in the color term embedding and perceptual color spaces, where $N$ is the number of color terms considered. For our analysis, we exclude color terms used less frequently than a cutoff $f=100$ in the color lexicon, leaving us with the $18$ most commonly used color terms.\footnote{This includes all color terms which are considered "basic" (\textit{red}, \textit{blue}, etc.), and commonly used "derived" terms (\textit{pink}, \textit{gray}, \textit{turquoise}, \textit{maroon}, etc.), but excludes the rest which are only infrequently used as color terms (\textit{forest}, \textit{puke}, \textit{dew}, \textit{seafoam}, etc.). See appendix~\ref{app:color_terms} for full list of colors included.}  We then separately construct an $N \times N$ RSM for each of the LM spaces and for CIELAB . Each cell in the RSM corresponds to the similarity between the activity patterns associated with pairs of experimental conditions $n_i, n_j \in N$.

For the color term embedding space, we employ Pearson's correlation coefficient ($r$) as a similarity measure between each pair of embeddings $n_i,n_j \in N$. For the CIELAB space, we elect to use the following method, per \citepos{regier2007color} suggestion: $sim(n_i, n_j)=\exp(-c \times [dist(n_i, n_j)]^2)$, where $c$ is a scaling factor (set to $0.001$ in all experiments reported here) and $dist(n_i, n_j)$ is the CIELAB distance ($\Delta$ E\_{CMC}$^{*}$)\footnote{We use the \texttt{colormath} Python package, setting illuminant to C, and assuming 2 degree standard observer.} between chips $n_i$ and $n_j$. This similarity measure is derived from the psychological literature on categorization and is meant to model the assumption that beyond a certain distance colors appear entirely
different, so that increasing the distance has no further effect on dissimilarity. Finally, we report the mean Kendall's $\tau$ between the color term embedding and color space RSMs. We also report $\tau$ per color term (i.e. per row in the RSM), which corresponds to how well-aligned each individual color term is. 

\paragraph{Linear mapping}
\newcommand{\Lagr}{\mathcal{L}}
\newcommand{\norm}[1]{\left\lVert#1\right\rVert}

\newcommand\thetaf{\ensuremath{{\Theta_{jr}}}}
\newcommand\csubj{\ensuremath{X}}
\newcommand\cf{\ensuremath{Y}}
\newcommand\decoder{\ensuremath{W}}
\newcommand\decoderloss{\ensuremath{\Lagr(W;\alpha)}}

We train regularised linear regression models to map from color term embedding space $X \in \mathbb{R}^{n\times d_{LM}}$ to CIELAB space $Y \in \mathbb{R}^{n\times 3}$, minimising $\decoderloss = \norm{\csubj \decoder - \cf}_2^2 + \alpha \norm{\decoder}_1$, where $\decoder \in \mathbb{R}^{3 \times d_{LM}}$ is a linear map and $\alpha$ is the lasso regularization hyper parameter. We vary $\alpha$ across a wide range of settings to examine the effect of probe complexity, which we measure using the nuclear norm of the linear projection matrix  $W \in \mathbb{R}^{\phi \times \iota}$; $||W||_*=\sum_{i=1}^{min(\phi,\iota)} \sigma_i(W)$, where $\sigma_i(W)$ is the $i$th singular value of $W$ \cite{pimentel2020pareto}.
The fitness of the regressors, evaluated using $n$-fold cross-validation ($n = 6$) indicates the alignability of the two spaces, given a linear transformation. Centroids corresponding to each Munsell color chip are computed in the color term embedding space via the weighted mean of the embeddings of the 51 terms used to label it. As in the RSA experiments, terms occurring less frequently than the cutoff ($f=100$) are excluded. For evaluation, we compute the average (across splits and datapoints) proportion of explained variance as well as the ranking of a predicted color term embedding according to the Pearson distance ($1 - r$) to gold.

\begin{table*}[ht!]
    \centering
    \footnotesize
    \resizebox{\textwidth}{!}{\begin{tabular}{c|cccc|cccc|cccc}
    \toprule
    & \multicolumn{4}{c|}{NC} &\multicolumn{4}{c|}{RC} &\multicolumn{4}{c}{CC} \\
    \midrule
    \multirow{2}{*}{Model} & \multicolumn{2}{c}{RSA} & \multicolumn{2}{c|}{lin. map} &\multicolumn{2}{c}{RSA} & \multicolumn{2}{c|}{lin. map} & \multicolumn{2}{c}{RSA} & \multicolumn{2}{c}{lin. map} \\
    &max & mean &max & mean &max & mean &max & mean &max & mean &max & mean \\
    \midrule
    BERT &0.16$^*$ &0.01$_{\pm0.09}$ &0.75 &0.73$_{\pm0.01}$ &0.26$^\dagger$ &0.20$_{\pm0.03}$ &0.74 &0.73$_{\pm0.08}$ &0.24$^\dagger$ &0.19$_{\pm0.03}$ &0.76 &0.75$_{\pm0.05}$ \\
    RoBERTa &0.33$^\S$ &0.02$_{\pm0.11}$ &0.75 &0.73$_{\pm0.01}$ &0.20$^*$ &0.14$_{\pm0.04}$ &0.74 &0.73$_{\pm0.01}$ &0.19$^*$ &0.14$_{\pm0.04}$ &0.77 &0.76$_{\pm0.09}$ \\
    ELECTRA &0.13$^{\phantom{\S}}$ &0.01$_{\pm0.08}$ &0.75 &0.64$_{\pm0.13}$ &0.25$^\dagger$ &0.19$_{\pm0.05}$ &0.75 &0.73$_{\pm0.01}$ &0.23$^\dagger$ &0.16$_{\pm0.04}$ &0.78 &0.76$_{\pm0.01}$ \\
    \bottomrule
    \end{tabular}}
    \caption{Results for the RSA experiments show max and mean (across layers) Kendall's $\tau$;  correlations that are significantly non-zero are marked with *, $\dagger$ and $\S$ for $p < 0.05$, $< 0.01$ and $< 0.001$ respectively.  Results for the linear mapping experiments show max and mean selectivity.}
    \label{tab:results}
\end{table*}

\paragraph{Control task}
As proposed by \newcite{hewitt2019designing}, we construct a random control task for the linear mapping experiments, wherein we randomly swap each color chip's CIELAB code for another. This is meant to break the mapping between the color chips and their corresponding terms. Control task results are reported as the mean of 10 different random re-mappings.  We report probe \textit{selectivity}, which is defined as the difference between proportion of explained variance in the standard experimental condition and in the control task \cite{hewitt2019designing}. We run similar control for the RSA experiments, where the CIELAB space centroids are randomly shuffled.

\section{Results}

\begin{figure*}[ht]
\centering
\includegraphics[width=\textwidth]{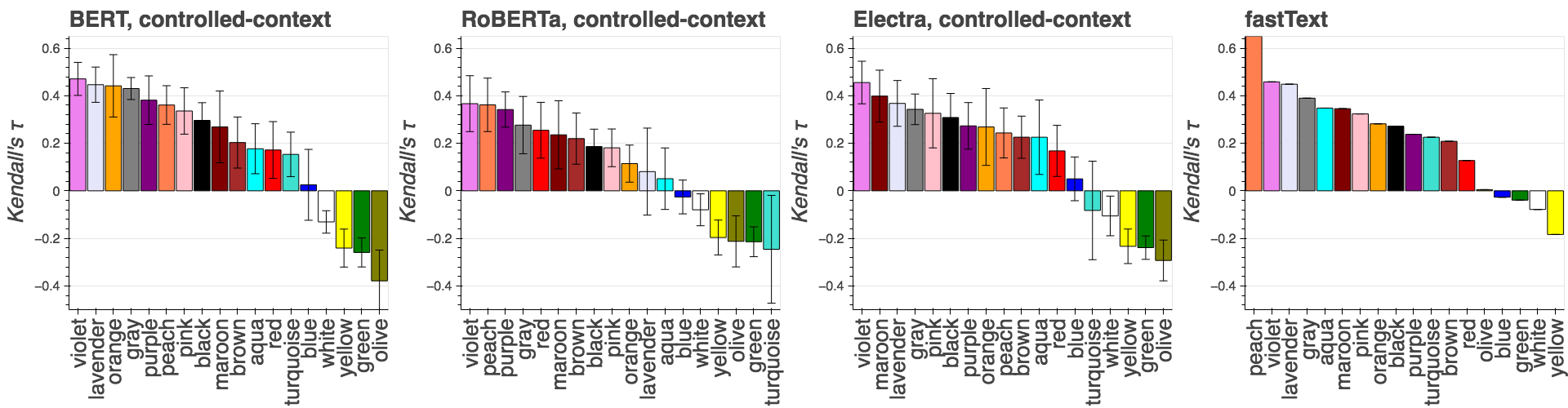}
\caption{RSA results (Kendal's $\tau$) broken down by color term for each of the LMs under the CC configuration and for the fastText baseline.}
\label{fig:rsa_per_color}
\end{figure*}

 Table ~\ref{tab:results} shows the max, mean, and standard deviation (across layers) of alignment scores for each of the LMs, per alignment method and setting. For RSA, we observe significant correlations across all configurations: most LM layers show a topological alignment with color space. Notably, this is also true for the static embeddings and for one of the PMI baselines (Table \ref{tab:base_results}). Although some variance is observed,\footnote{In particular, results for NC show large variances across layers. The mean correlation across layers in this setup is near zero, even though max correlations for BERT and RoBERTa are significant; this is unsurprising, however, as the LM has likely never encountered single color term tokens in isolation (cf. \citet{bommasani-etal-2020-interpreting})} the presence of significant correlations is telling, given the small sample size (18). Furthermore, randomly permuting the color space centroids leads to RSA correlations that are non-significant for all setups ($p > 0.05$), which lends further credence to models' alignment with CIELAB structure.  
 
 Figure~\ref{fig:rsa_per_color} shows the breakdown of correlations per color term for the three LMs under CC, as well as for fastText. We find that this ranking of color terms is largely stable across models and layer. Full RSMs for all models and CIELAB are in appendix~\ref{app:rsms}. The RSMs show evidence of the higher correlations for colors like violet, orange, and purple, being driven by general clusterings of similarity/dissimilarity. For instance, for both the CIELAB and CC BERT RSMs, violet's top \textit{nearest} neighbors include purple, lavender, pink, and orange, and its \textit{furthest} neighbors include aqua, olive, black, and gray. Correlations do not, however, appear to be driven by consistently aligned partial orderings within the clusters. In addition, we compute RSA correlations between the different models. Results show that NC embeddings have low alignment to all others (details in appendix \ref{app:model_rsa}).

For the linear mapping experiments, we observe the highest selectivity scores for CC (Table \ref{tab:results}, right) compared to NC and RC (Table \ref{tab:results}, left, middle) and baselines (Table \ref{tab:base_results}). This validates our intuition that controlling for variation in sentence context would reveal increased alignment to color space. 

\begin{figure}[ht]
 \begin{subfigure}{\columnwidth}
\centering
\includegraphics[scale=0.23]{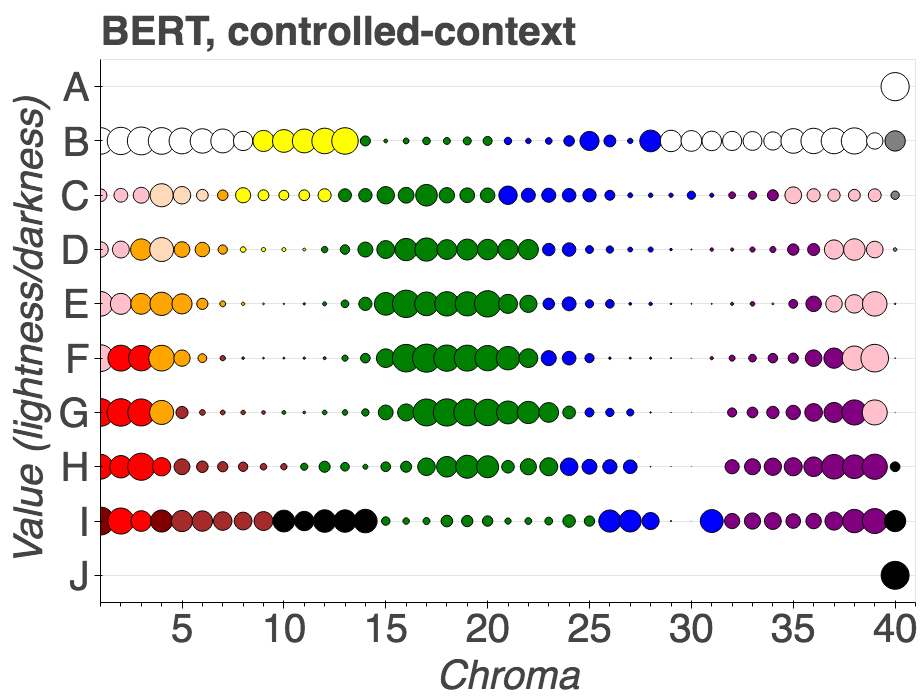}
\caption{Each circle on the chart represents the ranking of the predicted color chip when ranked according to Pearson distance from gold (larger circle $\cong$ higher/better ranking).}
\label{fig:munsell_bert}
\end{subfigure}
 \begin{subfigure}{\columnwidth}
\centering
\includegraphics[scale=0.23]{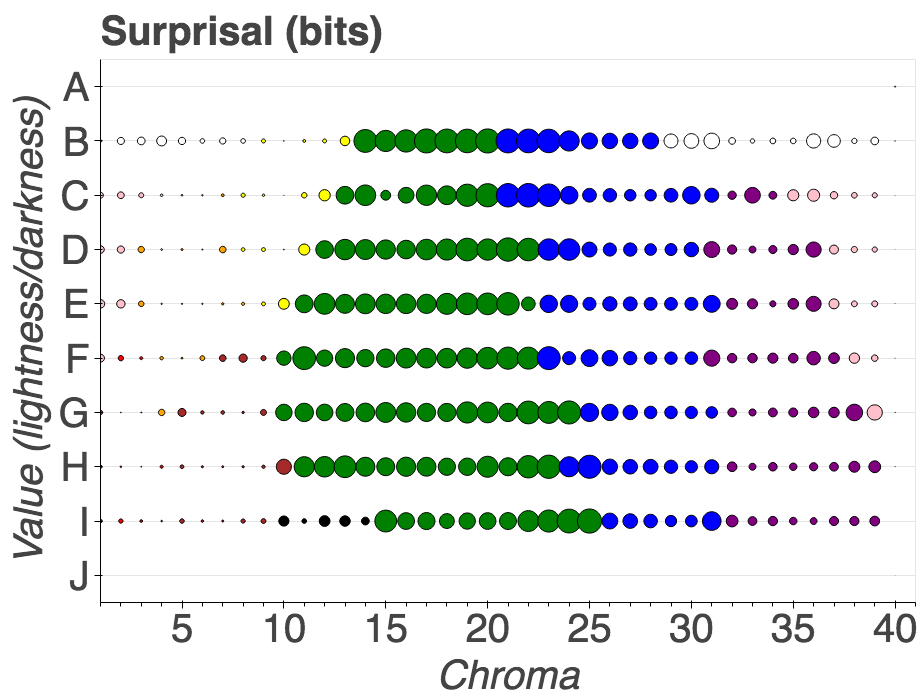}
\caption{Each circle on the chart represents a color chip's suprisal score (larger circle $\cong$ higher score).}
\label{fig:surprisal}
\end{subfigure}
\caption{(a) shows linear mapping results for BERT, under the CC configuration, broken down by Munsell color chip; (b) shows suprisal per chip. Circle colors reflect the modal color term assigned to the chips.}
\end{figure}

Furthermore, we observe that, over the full range of probe complexities for the experimental condition and the control task (described as in \S\ref{sec:eval}), all models demonstrate high selectivity (see \ref{app:complexity} for full results). It is, therefore, safe to attribute the fitness of the probes to information encoded in the color term representations, rather than to memorization. In terms of individual colors, Figure~\ref{fig:munsell_bert} depicts the ranking of predicted CIELAB codes per Munsell color chip for BERT (CC). We find that these results are largely stable across models and layers (see appendix \ref{app:munsell_full} for full set of results and for reference chart). Also, we observe that clusterings of chips with certain modal color terms (\textit{green, blue}) show worse rankings than the rest.

\begin{table}[t!]
{\small
    \centering
\begin{tabular}{c|cc}
        \toprule
        Model & RSA  & lin. map \\
        \midrule
        pmi-1 & 0.14$^{\phantom{*}}$ & 0.72 \\
        pmi-2 & 0.11$^{\phantom{*}}$ &  0.70 \\
        pmi-3 & 0.17$^*$ & 0.71\\
        fastText & 0.23$^*$ & 0.72 \\
        \midrule
\end{tabular}
 \caption{Baseline results. RSA results show Kendall's $\tau$; results with * are significantly non-zero ($p < 0.05$). Linear mapping results show selectivity.}
\label{tab:base_results}
}
\end{table}

\section{Analysis and Discussion}

Having demonstrated the existence of models'  alignment to CIELAB across various configurations, we now present an analysis and discussion of these results.

\paragraph{Dimensionality of color subspace}
Previous work has shown that linguistic information such as part-of-speech category, dependency relation type, and word sense, is expressed in low-dimensional subspaces of language model representations \cite{reif2019visualizing, durrani-etal-2020-analyzing, hernandez2021low}. We investigate the dimensionality of the subspace required to predict the CIELAB chip codes from the term embeddings, following the methodology of \newcite{durrani-etal-2020-analyzing}. Averaging over the three predicted CIELAB dimensions, we rank the linear mapping coefficients (from the experiments described in \S\ref{sec:method}), sorting the weights by their
absolute values in descending order. Results (appendix \ref{app:dims}) show that across models and layers, ${\sim}0.4$ of the variance in the CIELAB chip codes can be explained by assigning $95\%$ of the weights to ${\sim}10$ dimensions. $\numrange{30}{40}$ dimensions are sufficient to explain ${\sim}0.7$ of the variance, nearly the proportion of variance explained by the full representations (Table \ref{tab:results}).

\begin{table}[ht]
    \centering
    \footnotesize
     \resizebox{\columnwidth}{!}{%
    \begin{tabular}{ccc|cc}
        \toprule
        Model & RSA max & RSA mean & lin. map.. max & lin. map. mean \\
                \midrule
        BERT-mini & 0.077  & 0.043 $\pm$ 0.340 & 0.729 & 0.582 $\pm$ 0.291 \\
        BERT-small & 0.106  & 0.070 $\pm$ 0.191 & 0.734 & 0.598 $\pm$ 0.294 \\
        BERT-medium & 0.097  & 0.057 $\pm$ 0.035 & 0.739 & 0.654 $\pm$ 0.221 \\
        BERT-base & 0.162$^{*}$  & 0.092 $\pm$ 0.058 & 0.740 & 0.677 $\pm$ 0.182 \\

        \bottomrule
        
    \end{tabular}
    }
    \caption{Results for the four smaller BERT models. RSA results (left) show max and mean (across layers) Kendall's correlation coefficient ($\tau$). Correlations that are significantly non-zero are indicated with: * : $p < 0.05$.  Results for the Linear Mapping experiments (right) show max and mean selectivity. Standard deviation across layers is included with the mean results.}
    \label{tab:small_results}
\end{table}

\paragraph{Effect of model size} We also evaluate the effect of model size on alignment by testing four smaller BERT (CC) models\footnote{for details see appendix \ref{app:smaller}} using the same setup described above. The results (table \ref{tab:small_results}) show that alignment as measured by both RSA and linear mapping progressively increases with model size, meaning that that with growing complexity, model representational geometry of color terms moves towards isomorphism to CIELAB.

\paragraph{Color temperature}
In Figures \ref{fig:rsa_per_color} \& \ref{fig:munsell_bert} we observe that on average, warmer colors (\textit{yellow, orange, red, etc.}) show a closer alignment than cooler ones (\textit{blue, green, etc.}). In recent work, \newcite{gibson2017color} reported that the former are on average communicated more efficiently (see next paragraph) than the latter, across languages. This is attributed to warmer colors being more prevalent as colors of behaviorally relevant items in the environment --- salient objects --- compared to cooler ones, which occur more often as background colors. To verify this observation, we partition the space of chips into two (see appendix~\ref{app:temp} for details) and compute the average explained variance across warm and cool colors. The results (see appendix~\ref{app:temp} for plots) show that, term embeddings of warm colors are better aligned to CIELAB than those of cool ones, across models and configurations. This is consistent with the bias described in \newcite{gibson2017color}, which we conjecture might be filtering through into the distributional statistics of (color terms in) textual corpora, influencing the representations learned by various methods which leverage these statistics.

\paragraph{Connection to listener surprisal}
\newcite{gibson2017color}'s findings are based on the application of an information theoretic analysis to color naming, framing it as a communication game where a speaker has a particular color chip $c$ in mind and uses a word $w$ to indicate it then a listener has to correctly guess $c$, given $w$. Communication efficiency is measured through surprisal,  $S$, which in this setting corresponds to the average number of guesses an optimal listener takes to arrive at the correct color chip. We calculate $S(c)$ for each chip in the color lexicon. Surprisal is defined as $S(c)=\sum_{w} P(w|c) \cdot \log\left(\frac{1}{P(c|w)}\right)$, where $P(w|c)$ is the probability that a color c gets labeled as $w$ and $P(c|w)$ is computed using Bayes Theorem.
Here, $P(w)$ represents how often a particular word gets used across the color space (and participants), and $P(c)$ is a uniform prior. Figure \ref{fig:surprisal} shows surprisal per chip. High surprisal chips correspond to a lower color naming consensus among speakers, meaning that a more variable range of terms is used for these (color) contexts. We hypothesize that this could be reflected in the representations of color terms corresponding to high surprisal chips. To test this, we compute Spearman's correlation ($\rho$) between a chip's regression score (predicted color chip code ranking) and its surprisal. We find significant Spearman's rank correlation between lower ranking and higher surprisal for all LMs under all configurations ($0.12 \leq \rho \leq 0.17$, $p < 0.05$). 

\paragraph{What factors predict color space alignment?}
Given that LMs are trained exclusively on text corpora, we hypothesize that alignment between their embeddings and CIELAB is influenced by corpus usage statistics. To determine which factors could predict alignment score, we extract color term log frequency, part-of-speech tag (POS), dependency relation (DREL), and dependency tree head (HEAD) statistics for all color terms from a dependency-parsed \cite{straka2016udpipe} common crawl corpus. In addition to this, we compute, per color term, the entropy of its normalised PMI distribution  (\texttt{pmi-col}, see \S\ref{sec:method}) as a measure of collocation.\footnote{Low entropy reflects frequent co-occurrence with a small subset of the vocabulary and high entropy the converse.} 
We then fit a Linear Mixed Effects Model \cite{galecki2013linear} to the features listed above, with RSA score (Table \ref{tab:results}) as the response variable, and model type as a random effect. 

We follow a multi-level step-wise model building sequence, where a baseline model is first fit with color term log frequency as a single fixed effect. 
A model which includes \texttt{pmi-col} as an additional fixed effect is then fit, and these two terms are included as control predictors in all later models. Following this, we compute POS, DREL, and HEAD lemma distribution entropies per color term (\texttt{pos-ent, deprel-ent, head-ent}). Higher entropies indicate that the term is employed in more diverse contexts with respect to those categories. Following entropy computation, we separately fit models including each three entropy statistic features. Finally, we calculate the proportion of: POS tags that are adjectives, \texttt{adj-prop}; DRELs that are adjectival modifiers, \texttt{amod-prop}; and those that are copulas, \texttt{cop-prop}. The first two evaluate the effect of a color term occurring more or less often as an adjectival modifier, while the latter tests the hypothesis that assertions such as \textit{The banana is yellow} could provide indirect grounding \cite{merrill2021provable}, thereby leading to higher alignment. Including the entropy term which led to the best fit (\texttt{deprel-ent}) in the previous level, models are fit including terms for each of the proportion statistics. Model comparison is carried out by computing the log likelihood ratio between models that differ in a single term. See appendix \ref{app:lme} for model details. 

Results show that: 

\begin{itemize}
    
\item  \texttt{pmi-col} significantly improves fit above log frequency and has a negative coefficient,  meaning that terms that occur in more fixed collocations are less aligned to the perceptual space. Intuitively, this makes sense as the color terms in many collocations such as e.g. \textit{Red Army} or \textit{Black Death} are employed in contexts which are largely metaphorical rather than attributive or descriptive. 

\item \texttt{deprel-ent} and \texttt{head-ent} (but not \texttt{pos-ent}) lead to a significantly improved fit compared to the control predictors; we observe positive coefficients for both, indicating RSA score is higher for terms that occur in more varied syntactic dependency relations and modify a more diverse set of syntactic heads. This suggests that occurring in a more diverse set of contexts might be beneficial for robust representation learning, in correspondence with the idea of sample diversity in the active learning literature \cite{brinker2003incorporating,yang2015multi}. \texttt{pos-ent}'s lack of  significance, on the other hand, indicates that the degree of specification offered by the POS tagset might be too coarse to meaningfully differentiate between color terms, e.g. nouns can occur in a variety of DRELs such as subjects, objects, oblique modifiers (per the Universal Dependecies \cite{nivre-etal-2020-universal}).  

\item out of the proportion statistics, only the \texttt{amod-prop} term improves fit; it has a positive coefficient, thus color terms occurring more frequently as adjectival modifiers show higher scores. \texttt{adj-prop} is not significant, providing further evidence for the POS tagset's level of granularity being too coarse. Finally, as \texttt{cop-prop} is not significant, it appears that occurring more frequently in assertion-like copula constructions does not confer an advantage in terms of alignment to perceptual structure.

\end{itemize}

\paragraph{Vision-and-Language models}
In a preliminary set of experiments, we evaluated multi-modal Vision-and-Language models (VisualBERT \cite{li2019visualbert} and VideoBERT \cite{sun2019videobert}), finding no major differences in results from the text-only models presented in this study.  

\section{Related Work}
Distributional word representations have long been theorized to capture various types of information about the world \cite{schutze1992dimensions}. Early work in this regard employed semantic similarity and relatedness datasets to measure alignment to human judgements \cite{agirre2009study, bruni2012distributional, hill2015simlex}. \citet{rubinstein2015well}, however, question whether the distributional hypothesis is equally applicable to all types of semantic information, finding that taxonomic properties (such as animacy) are better modelled than attributive ones (color, size, etc.). To a similar end, \citet{lucy2017distributional} analyze how well distributional representations encode various aspects of grounded meaning.
They investigate whether language models would \textit{``be worse off
for not having physically bumped into walls before they hold discussions on wall-collisions?''}, finding that perceptual features are poorly modelled compared to encyclopedic and taxonomic ones. 

More recently, several studies have asked related questions in the context of language models. For example, \citet{davison-etal-2019-commonsense} and \citet{petroni-etal-2019-language} mine LMs for factual and commonsense knowledge by converting knowledge base triplets into cloze statements that are used to query the models. In a similar vein, \citet{forbes2019neural} investigate LM representations' encoding of object properties (e.g., \textit{oranges are round}), and affordances (e.g. \textit{oranges can be eaten}), as well as the interplay between the two. \citet{weir2020probing} demonstrate that LMs can capture \textit{stereotypic tacit assumptions} about generic concepts, showing that they are adept at retrieving concepts given their associated properties (e.g., \textbf{bear} given \textit{A \_\_\_ has fur, is big, and has claws.}). Similar to other work, they find that LMs better model encyclopedic and functional properties than they do perceptual ones. 

In an investigation of whether or not LMs are able to overcome reporting bias, \citet{shwartz-choi-2020-neural} extract all sentences in Wikipedia where one of $11$ color terms modifies a noun and test how well predicted the color term is when it is masked. They find that LMs are able to model this relationship between concepts and associated colors to a certain extent, but are prone to over-generalization. Finally, \citet{ilharco2020probing} train a probe to map LM representations of textual captions to paired visual representations of image patches, in order to evaluate how useful the former are for discerning between different visual representations. They find that many recent LMs yield representations that are effective at retrieving semantically-aligned image patches, but still far under-perform humans.

\section{Outlook}
It is commonly held that the learning of phenomena which rely on sensory perception is only possible through direct experience. Indeed, the view that people born blind could not be expected to acquire coherent knowledge about colors has been prevalent since at least the empiricist philosophers  \cite{locke1847essay, hume1938abstract} and still holds currency  \cite{jackson1982epiphenomenal}.  Nevertheless, recent research highlighting the contribution of language and of semantic associations between concepts towards learning has demonstrated that the congenitally blind do in fact show a striking understanding of both color similarity \cite{saysani2018colour} and object colors \cite{kim2020shared}. 

This paper investigated whether representations of color terms that are derived from text only express a degree of isomorphism to the structure of humans' perceptual color space.\footnote{Clearly, complete isomorphism is rather unlikely: language in general, and color terms by extension, are far from being simply denotational, and language interacts with and is influenced by a myriad of factors besides perception.} Results from our experiments evidenced that such a topological correspondence exists. Notably, color term representations based on simple co-occurance statistics already demonstrated correspondence; those extracted from language models aligned more closely. We observed that warm colors, on average, show more alignment than cooler ones, linking to recent findings on communication efficiency in color naming \cite{gibson2017color}. 

Further analysis based on surprisal --- an information theoretic measure, used to evaluate how efficiently a color is communicated between a speaker and a listener --- revealed a correlation between lower topological alignment and higher color chip surprisal, suggesting that the kind of contexts a color occurs in play a role in determining alignment. Exploring this, we tested a set of color term corpus-derived statistics for how well they predict alignment, finding that a measure of a color term's collocationality corresponds to lower alignment, while the entropy of its dependency relation distribution and it occurring more frequently as and adjectival modifier correspond to closer alignment.

Our results and analyses present empirical evidence of topological alignment between text-based color term representations and perceptual color spaces. With respect to the debate started by \newcite{bender-koller-2020-climbing}, we hope that this work offers a modest step towards furthering our understanding of the kinds of ``meaning''  we expect language models to acquire, with and without grounded or embodied learning approaches, and that it will provide motivation for further work in this direction. 

\section*{Acknowledgements}
We would like to thank Vinit Ravishankar and Mitja Nikolaus for their feedback and comments. Mostafa Abdou and Anders Søgaard are supported by a
Google Focused Research Award and a Facebook
Research Award. 

\bibliography{anthology, custom}
\bibliographystyle{acl_natbib}

\appendix
\section{List of included color terms}
Red, green, maroon, brown, black, blue, purple, orange, pink, yellow, peach, white, gray, olive, turquoise, violet, lavender, and aqua. 
\label{app:color_terms}

\section{RSA between models}
\begin{figure*}[ht]
\centering
\includegraphics[width=\textwidth]{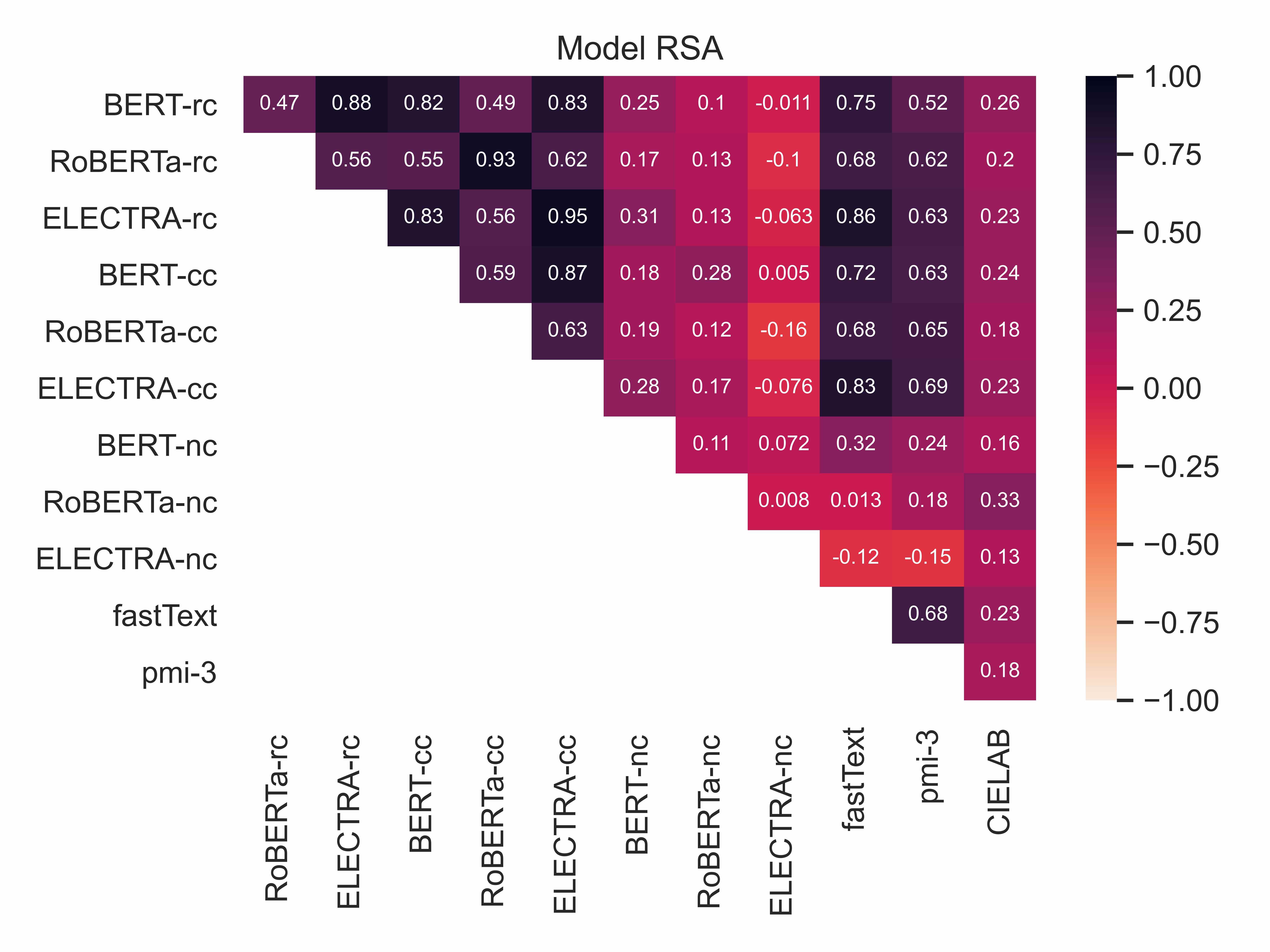}
\caption{Result of representation similarity analysis between all models (and configurations), showing Kendall's correlation coefficient between flattened RSMs. Results are shown for layers which are maximally correlated with CIELAB, per model. \texttt{-rc} indicates \textbf{random-context}, \texttt{-cc} indicates \textbf{controlled-context}, and \texttt{-nc} indicates \textbf{non-context}. }
\label{fig:rsa_heatmap}
\end{figure*}

Figure \ref{fig:rsa_heatmap} shows a the result of representation similarity analysis between the representations derived from all models (and configurations) as well as CIELAB, showing Kendall's correlation coefficient between flattened RSMs.
\label{app:model_rsa}

\section{Representation Similarity Matrices}

\begin{figure*}[ht]
\centering
\includegraphics[width=\textwidth]{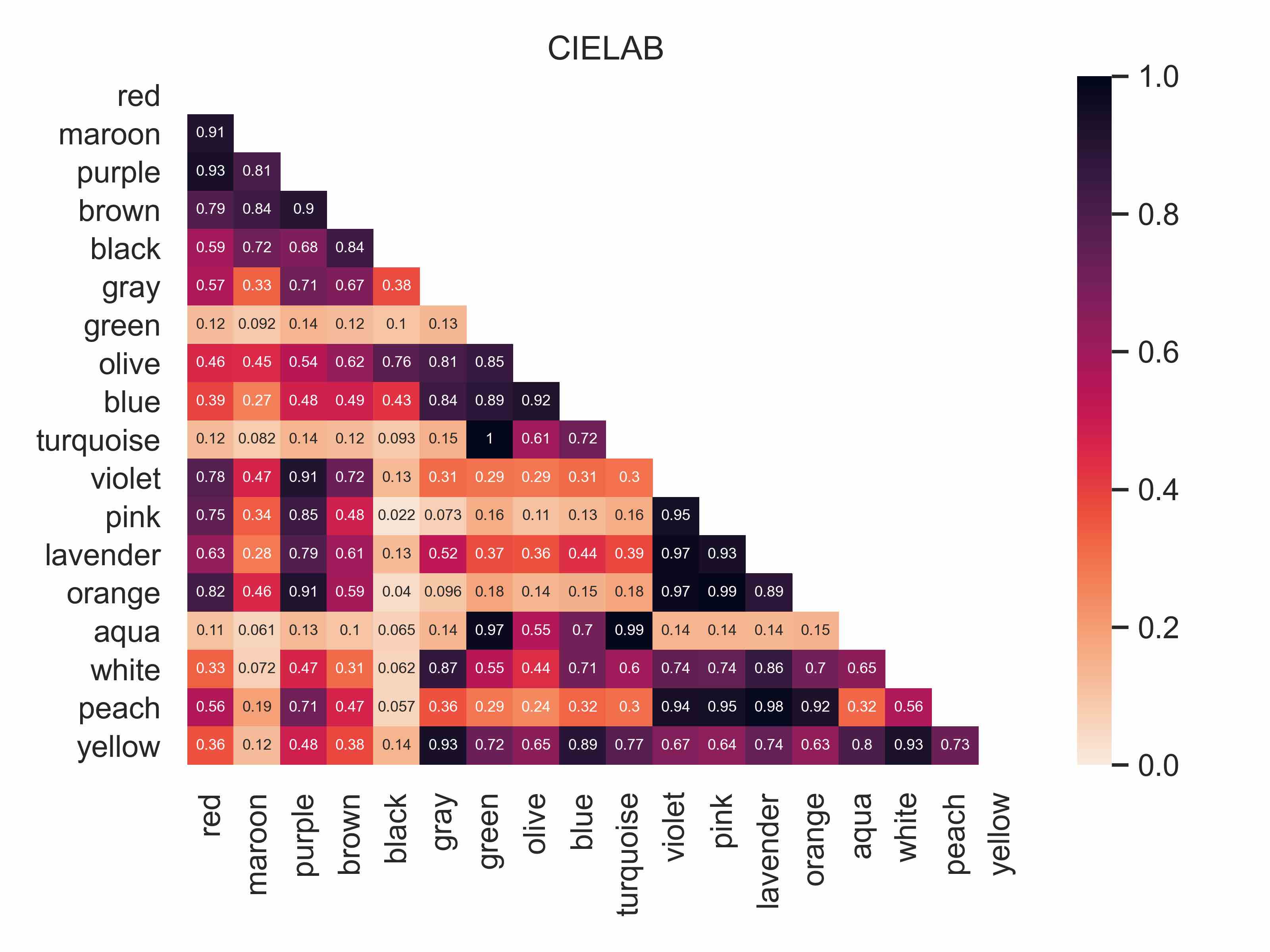}
\caption{CIELAB RSM}
\label{fig:cielab_rsm}
\end{figure*}

\begin{figure*}[ht]
\centering
\includegraphics[scale=0.99]{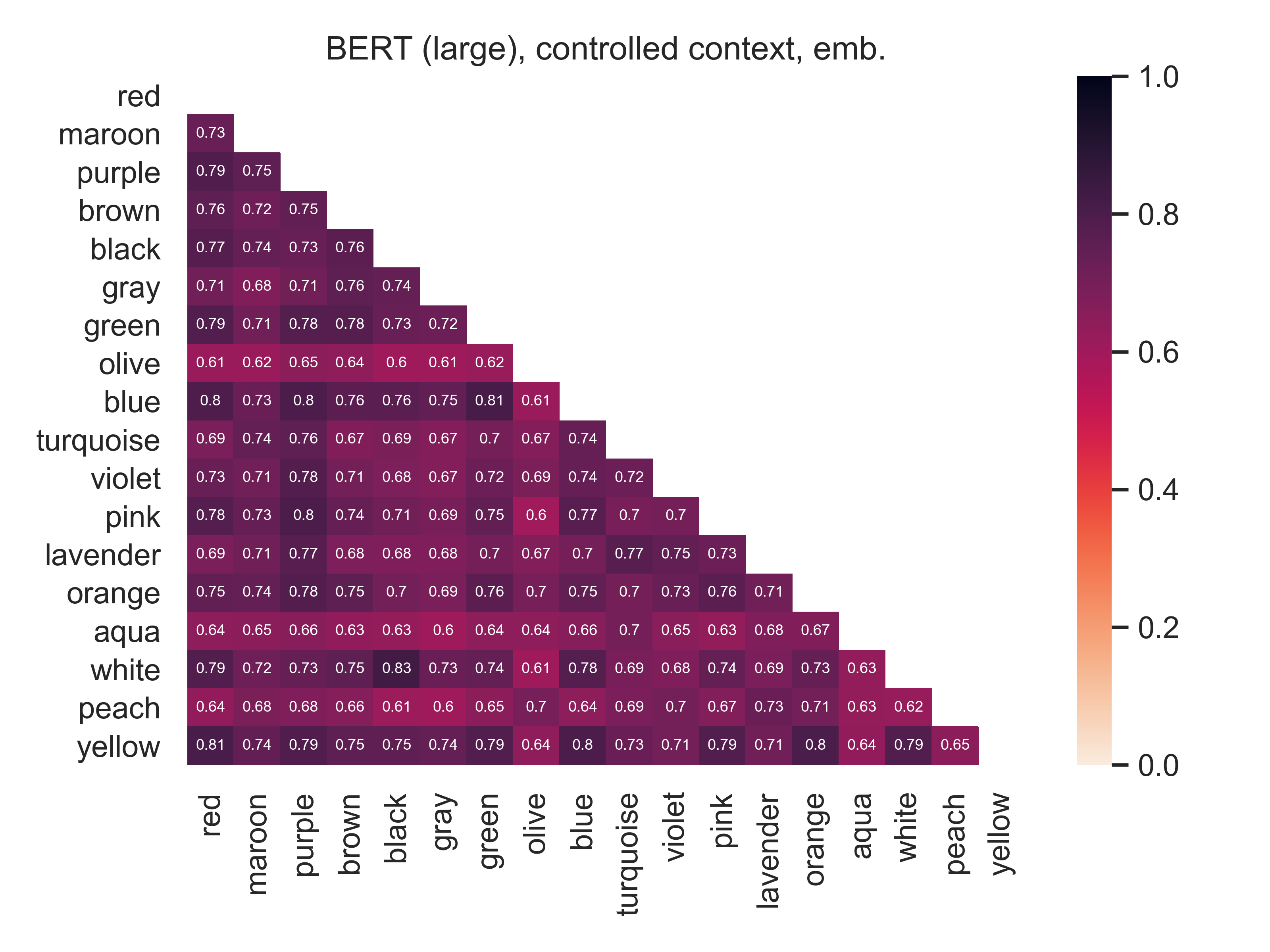}
\caption{BERT(CC) RSM}
\label{fig:bert_rsm}
\end{figure*}

\begin{figure*}[ht]
\centering
\includegraphics[width=\textwidth]{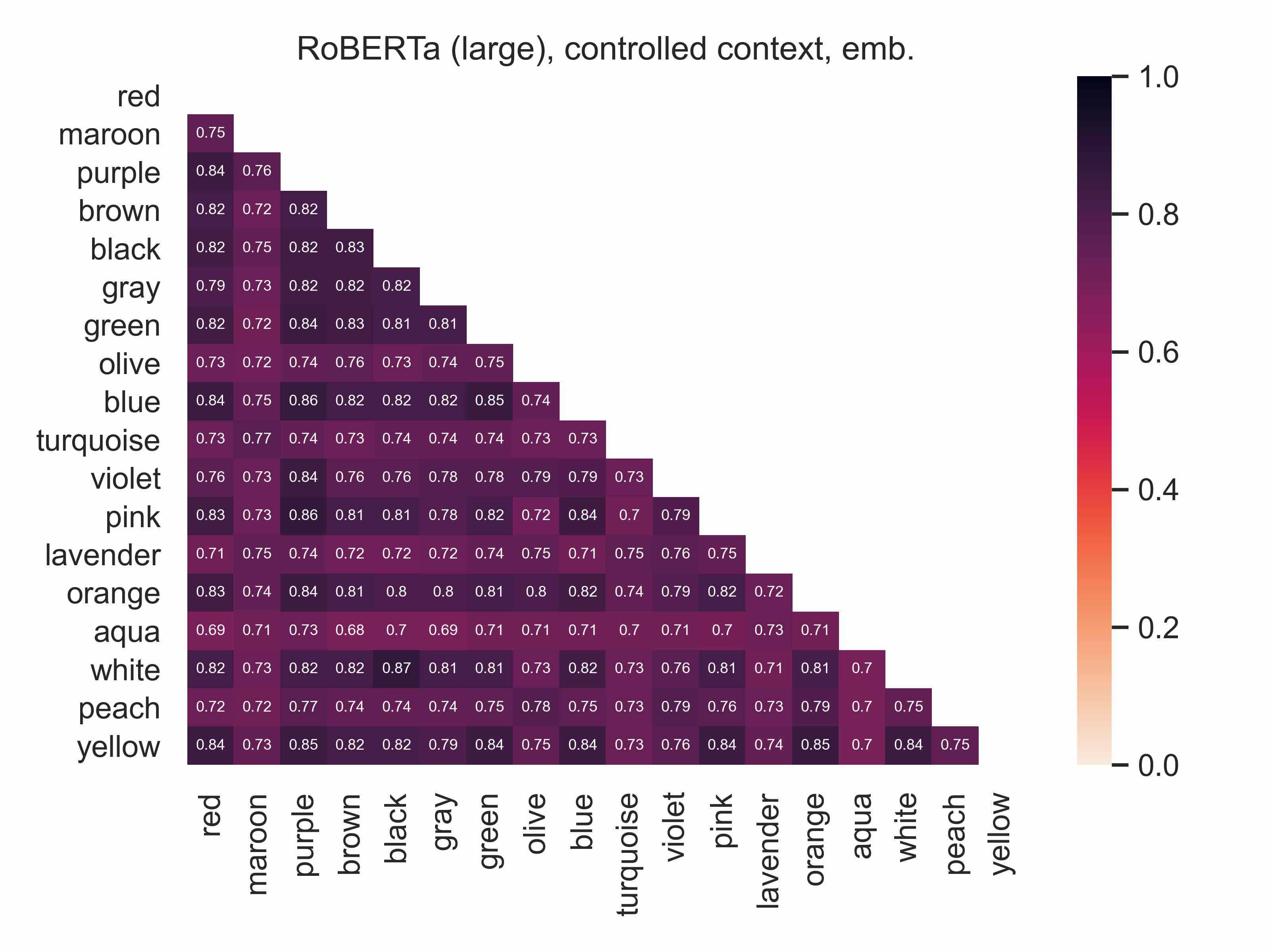}
\caption{RoBERTa(CC) RSM}
\label{fig:roberta_rsm}
\end{figure*}

\begin{figure*}[ht]
\centering
\includegraphics[width=\textwidth]{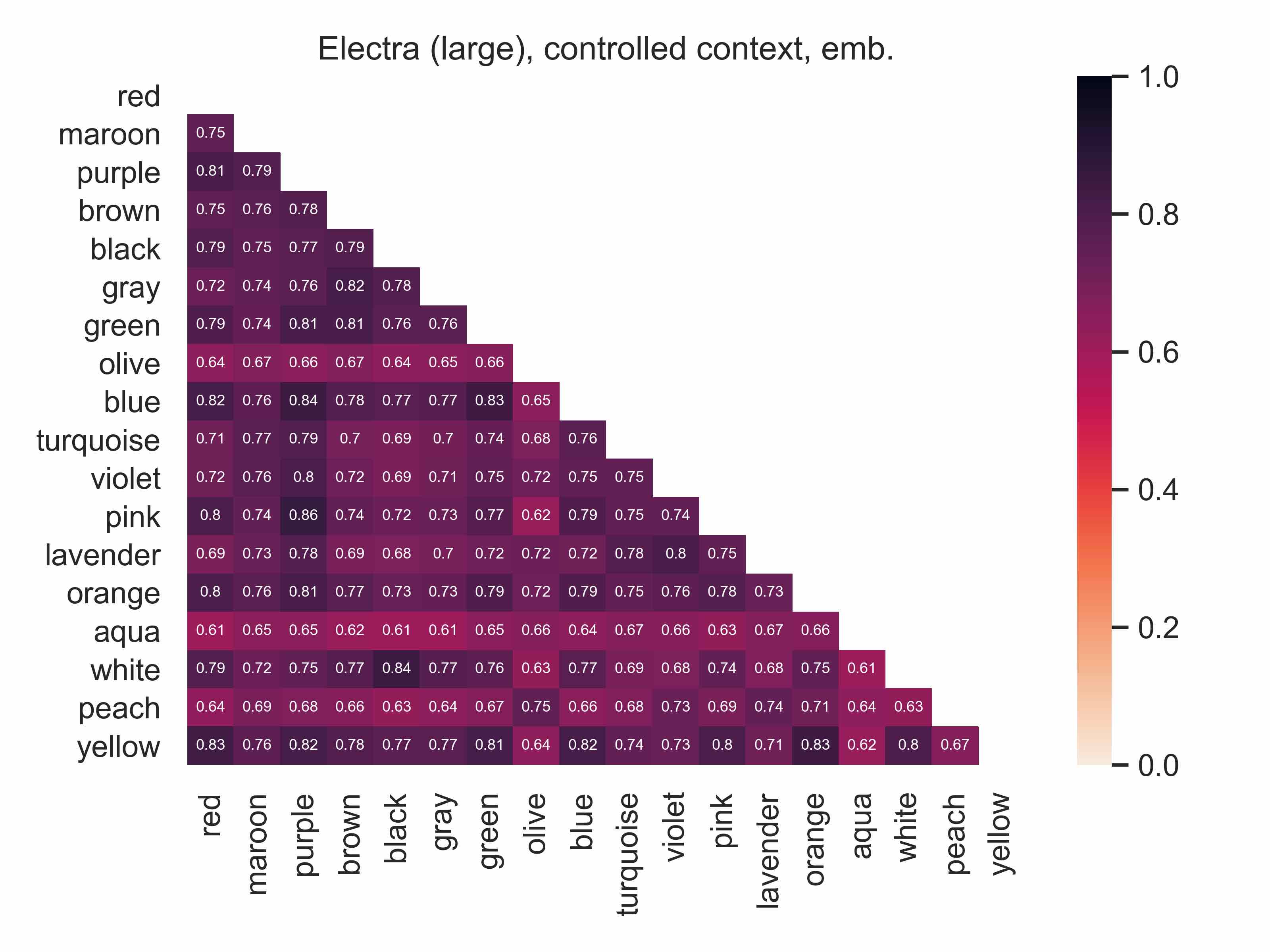}
\caption{ELECTRA(CC) RSM}
\label{fig:electra_rsm}
\end{figure*}

Figures \ref{fig:cielab_rsm} to \ref{fig:electra_rsm} show the representation similarity matrices employed for the RSA analyses, for the layer with the highest RSA score from each of the controlled-context (CC) models.

\label{app:rsms}

\section{Warm vs. Cool colors}
\label{app:temp}
\begin{figure*}[ht]
\centering
\includegraphics[scale=0.24]{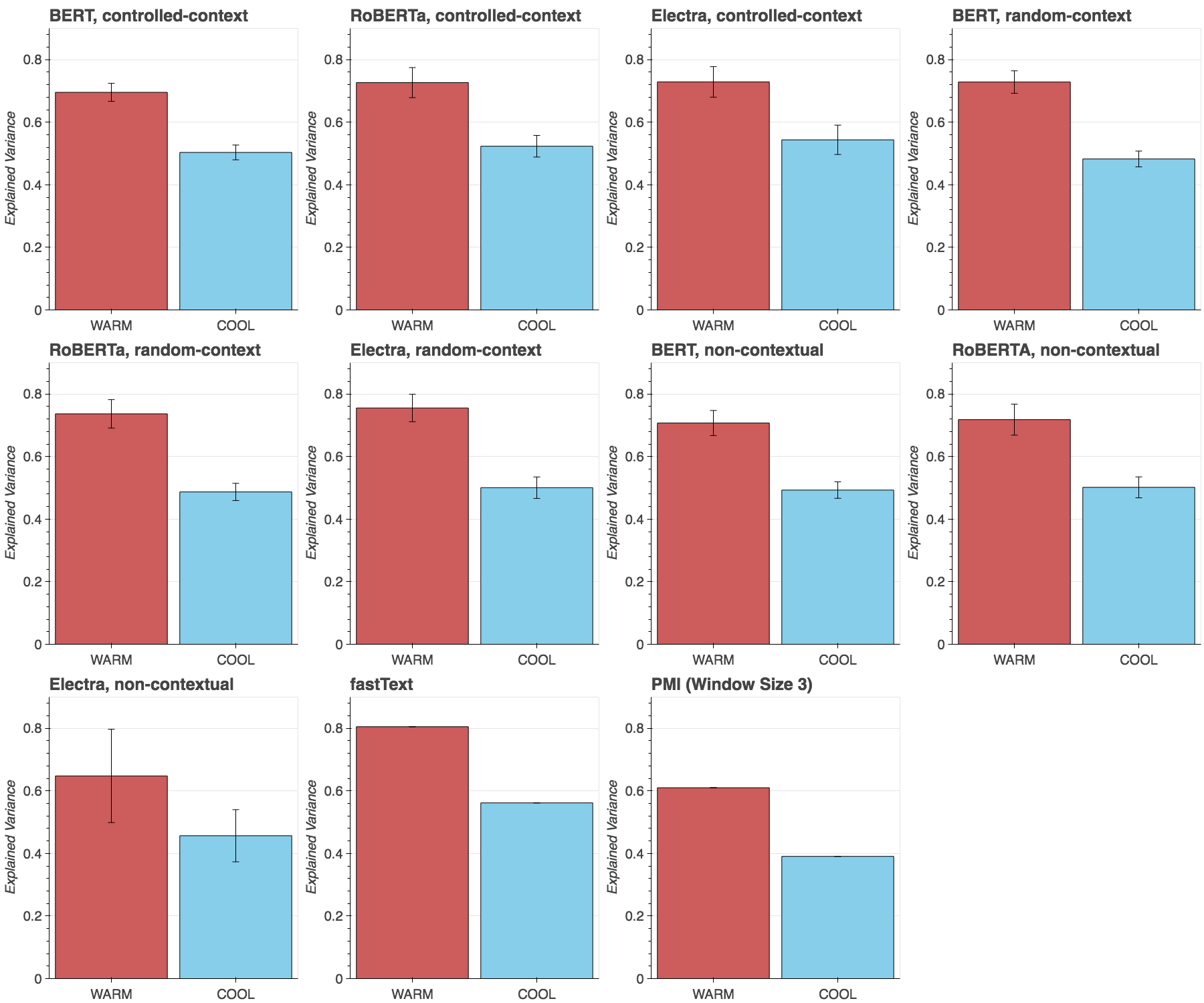}
\caption{Linear mapping results (proportion of explained variance) broken down by color chip temperature for each of the baselines and the LMs.}
\label{fig:temp_lm}
\end{figure*}

\begin{figure*}[ht]
\centering
\includegraphics[scale=0.24]{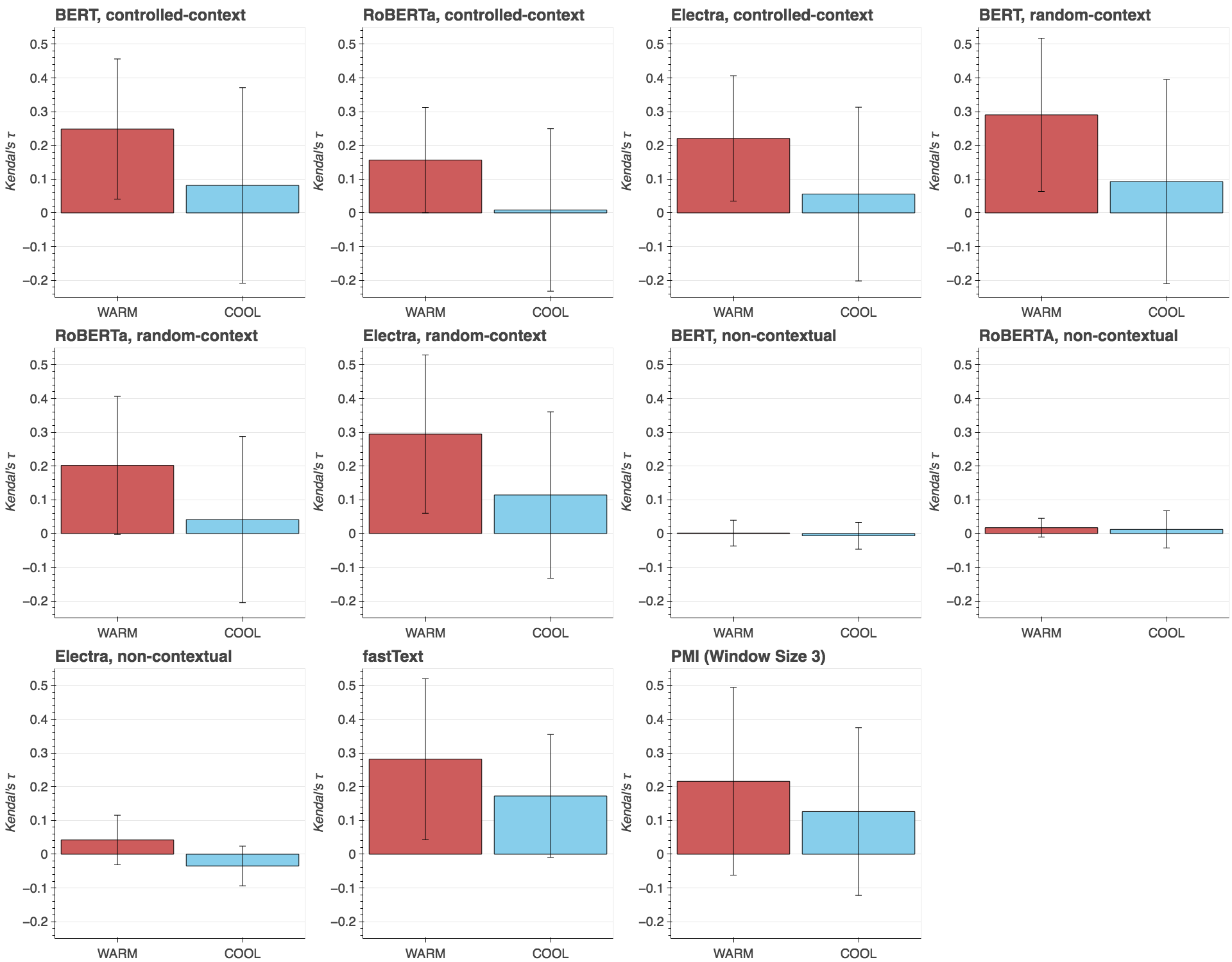}
\caption{RSA results (Kendall's $\tau$ ) broken down by color temperature for each for each of the baselines and the LMs.}
\label{fig:temp_rsa}
\end{figure*}

Figures \ref{fig:temp_lm} and \ref{fig:temp_rsa} show Linear Mapping and RSA results broken down by color temperature. The color space is split according to temperature measured according to the Hue dimension in the Hue-Value-Saturation space\footnote{\url{https://psychology.wikia.org/wiki/HSV\_color\_space}}.  

\section{Corpus statistics}
\begin{figure*}[ht]
\centering
\includegraphics[width=0.6\textwidth]{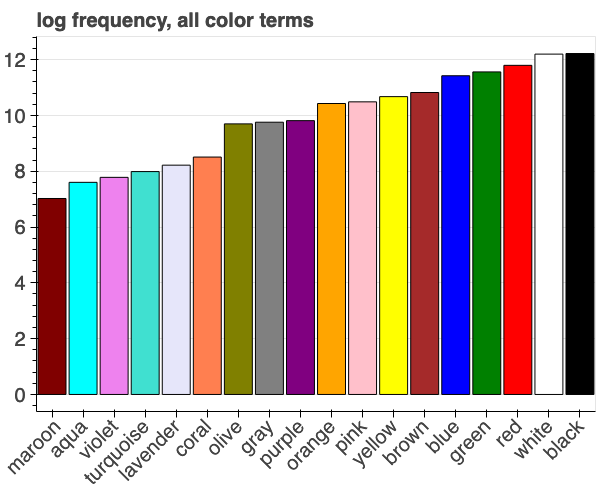}
\caption{Log frequency of color terms in common crawl.}
\label{fig:freq}
\end{figure*}

\begin{figure*}[ht]
\centering
\includegraphics[width=\textwidth]{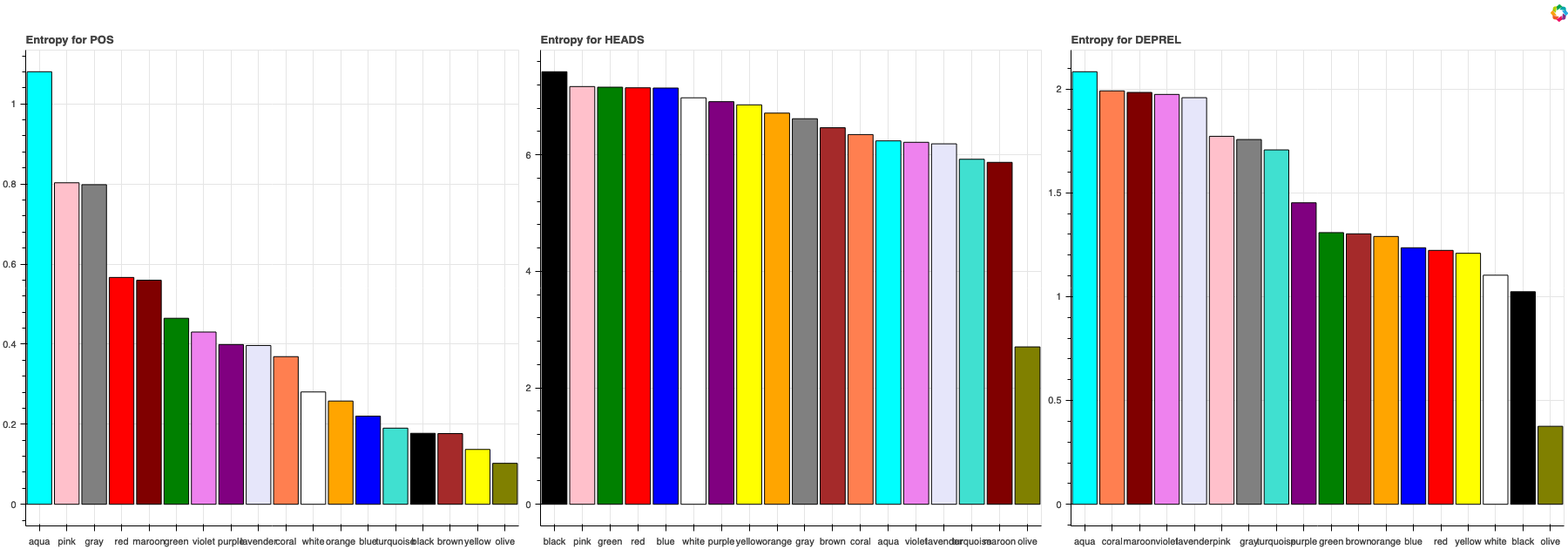}
\caption{Entropy of distributions over part-of-speech categories, dependency relations, and lemmas of dependency tree heads of color terms in common crawl.}
\label{fig:entropies}
\end{figure*}

Figures \ref{fig:freq} and \ref{fig:entropies} show log frequency and entropy of distributions over part-of-speech categories, dependency relations, and lemmas of dependency tree heads of color terms in common crawl.
 \label{app:corpus_stats}

\section{Linear mapping results by munsell color chip}
\begin{figure*}[ht]
\centering
\includegraphics[scale=0.16]{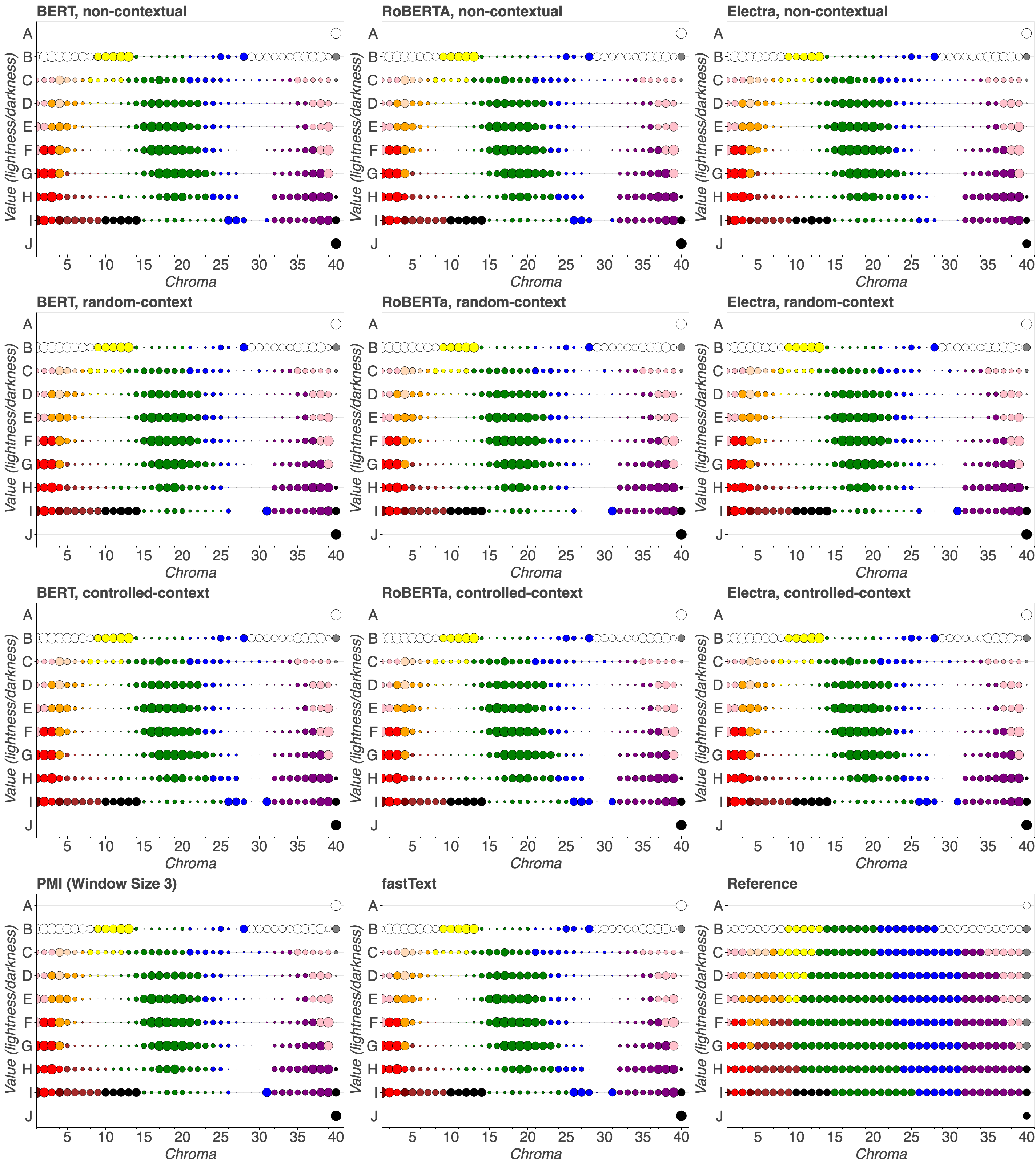}
\caption{Linear mapping results for each of the baselines and language models, under all extraction configurations, broken down by Munsell color chip. Each circle on the chart represents the ranking of the predicted color chip when ranked according to Pearson distance ($1 -$ Pearson's $r$) from gold -- the larger the circle, the higher (better) the ranking. Circle colors reflect the modal color term assigned to the chips in the lexicon. Reference plot showing modal color of all chips also included.}
\label{fig:munsell_full}
\end{figure*}

Figure \ref{fig:munsell_full} shows linear mapping results broken down by Munsell chip for all models and configurations.
\label{app:munsell_full}

\section{Linear mapping control task and probe complexity}

\label{app:complexity}
\begin{figure*}[ht]
\centering
         \centering
         \includegraphics[width=\textwidth]{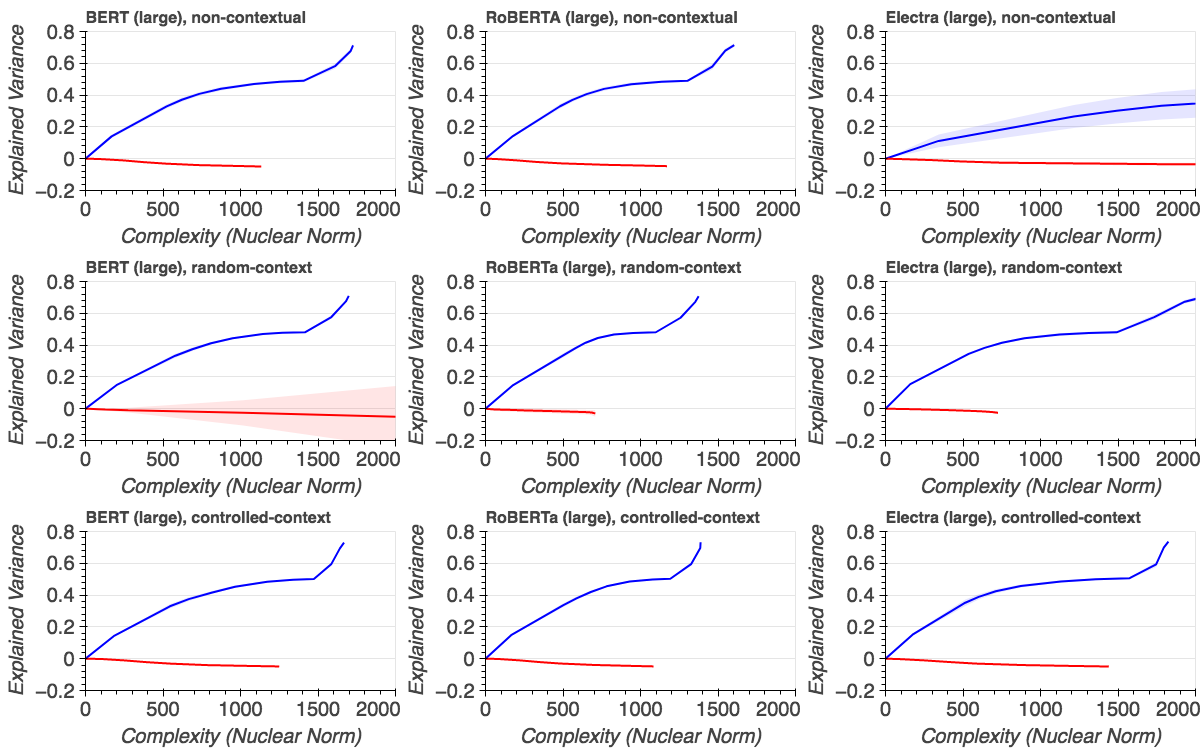}
\caption{Explained variance for the linear probes trained on the normal experimental condition (blue) and the control task (red) where color terms are randomly permuted. The means are indicated by the lines and standard deviation across layers is indicated by the bands.}
\label{fig:complexity}
\end{figure*}

Figure \ref{fig:complexity} shows the full results over a range of probe complexities for the standard experimental condition as well the random control task.

\section{Dimensionality of color subspace}
\label{app:dims}
\begin{figure*}[ht]
\centering
         \centering
         \includegraphics[width=\textwidth]{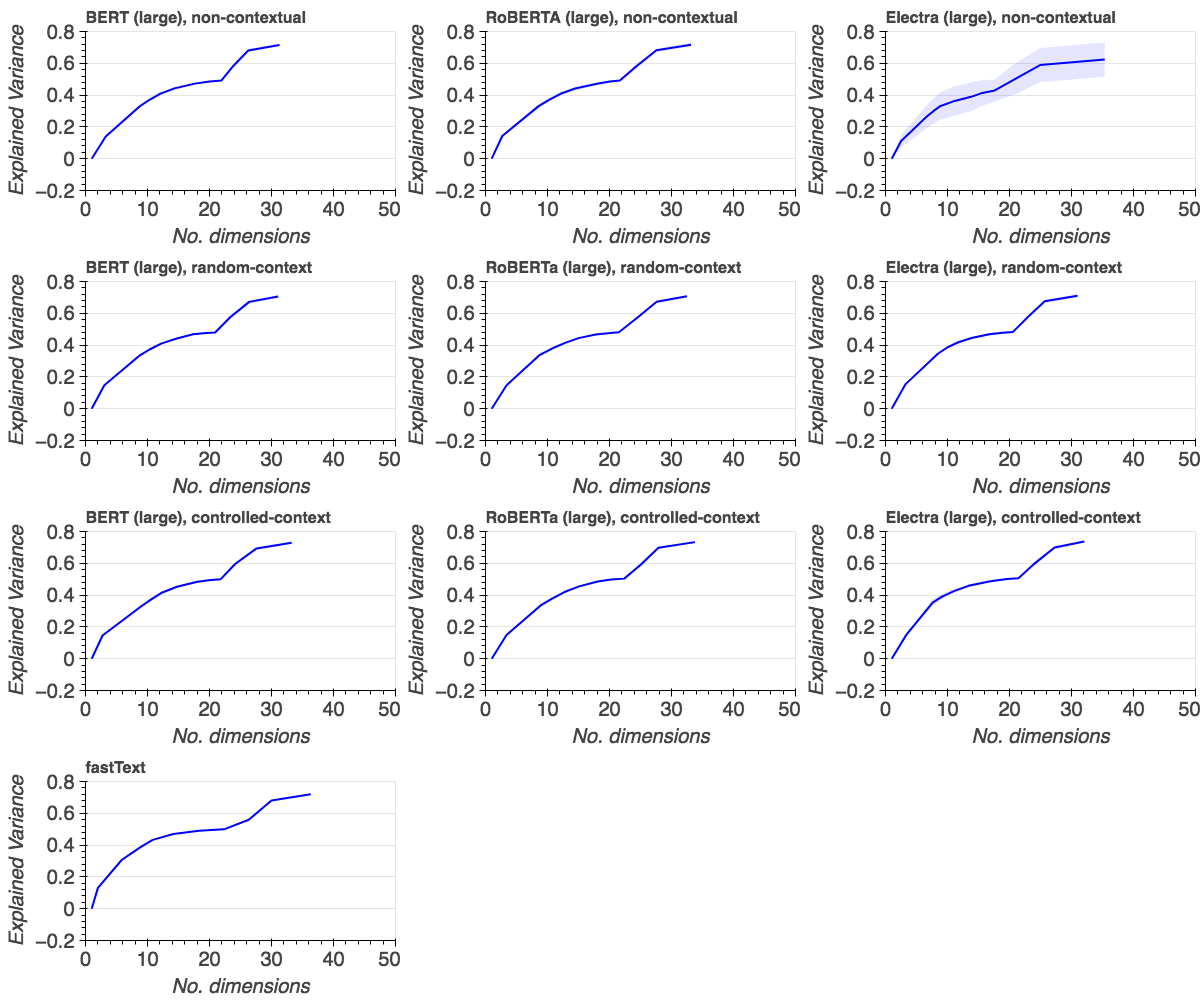}
\caption{The y-axis shows explained variance for the linear probes. The means are indicated by the lines and standard deviation across layers is indicated by the bands. The x-axis shows the number of regression matrix coefficients assigned $95\%$ of the weight.}
\label{fig:dims}
\end{figure*}

Figure \ref{fig:dims} shows the proportion of explained variance with respect to the number of dimensions which are assigned $95\%$ of the linear regression coefficient weights. 

\section{Effect of model size}
\label{app:smaller}
 Our model size experiments are run using four BERT models of different sizes: BERT-mini (4 layers, hidden size: 256), BERT-small (4 layers, hidden size: 512), BERT-medium (8 layers, hidden size: 512), and BERT-base (12 layers, hidden size: 768). Further model specification and training details for the first three can be found in \newcite{turc2019well} and for last in \newcite{devlin-etal-2019-bert}. 


        

\section{Linear Mixed Effects Model}
\label{app:lme}
To fit Linear Mixed Effects Models, we use the \textsc{LME4} package. With model type (BERT-CC, RoBERTa-NC, etc.) as a random effect, we follow a step-wise model construction sequence which proceeds along four levels of nesting: (i) in the first level color log-frequency is the only fixed effect, (ii) in the second \texttt{pmi-colloc} is added to that, (iii) in the third, each of \texttt{pos-ent, deprel-ent, head-ent} is added separately to the a model with log frequency and \texttt{pmi-colloc}, (iv) the term that leads to the best fit from the previous level \texttt{deprel-ent} is included, then each of the proportion terms \texttt{adj-prop, amod-prop, cop-prop} is added. The reported regression coefficients are extracted from the minimal model containing each term. 

\end{document}